\documentclass[11pt]{article}
\usepackage[a4paper,margin=1in]{geometry}
\usepackage{graphicx}
\usepackage{amsmath,amssymb}
\usepackage{hyperref}
\usepackage{xcolor}
\usepackage{listings}
\usepackage{authblk}
\usepackage{titlesec}
\usepackage{setspace}
\usepackage{enumitem}
\usepackage{microtype}
\usepackage{booktabs}
\usepackage{cleveref}
\usepackage{float}
\hypersetup{
  colorlinks=true,
  linkcolor=blue,
  citecolor=blue,
  urlcolor=blue,
  pdftitle={AI Scientific Assistant Core (AISAC): A Modular multi-agent Framework}, 
  pdfauthor={Chandrachur Bhattacharya, Sibendu Som},
  pdfkeywords={multi-agent systems, RAG, scientific AI, provenance, reproducibility}
}

\titleformat{\section}{\large\bfseries}{\thesection.}{0.5em}{}
\titleformat{\subsection}{\normalsize\bfseries}{\thesubsection}{0.5em}{}
\title{\textbf{AISAC: An Integrated Multi-Agent System for Transparent, Retrieval-Grounded Scientific Assistance}}

\author[1]{Chandrachur Bhattacharya\thanks{Email: cbhattacharya@anl.gov}}
\author[1]{Sibendu Som\thanks{Email: ssom@anl.gov}}
\affil[1]{Transportation and Power Systems Division, Argonne National Laboratory, USA}
\date{}

\begin{document}
\maketitle

\begin{abstract}
AI Scientific Assistant Core (AISAC) is a transparent, modular multi-agent runtime
developed at Argonne National Laboratory to support long-horizon,
evidence-grounded scientific reasoning and to serve as a practical daily tool
for domain scientists who require capable AI assistance without needing to
build or maintain AI infrastructure. Rather than proposing new agent
algorithms or claiming autonomous scientific discovery, AISAC contributes a
governed execution substrate that operationalizes key requirements for
deploying agentic AI in scientific practice, including explicit role semantics,
budgeted context management, traceable execution, and reproducible interaction
with tools and knowledge.

AISAC is structured around four design properties for scientific reasoning,
two of which are enforced as hard runtime invariants and two of which are
architectural objectives realized through system design:
(1) declarative agent registration with runtime-enforced role semantics and
automatic system prompt generation; (2) budgeted orchestration via explicit
per-turn context and delegation depth limits, with runtime interrupt and
clarification signals for human-in-the-loop oversight; (3) role-aligned memory access
across episodic, dialogue, and evidence layers; and (4) trace-driven
transparency through persistent execution records and a live event-stream
interface.

These properties are implemented through hybrid persistent memory (SQLite and
dual FAISS indices), governed retrieval with agent-scoped RAG, structured tool
execution with schema validation, and a configuration-driven bootstrap
mechanism that enables project-specific extension without modifying the shared
core. AISAC is in active use across multiple scientific workflows at
Argonne, including combustion science, materials research, and energy process
safety, illustrating its application as a reusable substrate for domain-specialized
AI scientific assistants across workstation, HPC, and restricted-network
environments. This manuscript presents the current architecture and design
rationale as a work in progress; comprehensive end-to-end evaluation across
scientific deployments is ongoing and will be reported in subsequent work.
\end{abstract}

\vspace{1em}

\section{Introduction}

Large language models (LLMs) have increasingly been explored as assistants for scientific research, motivated by their ability to synthesize information, generate code, and support complex reasoning tasks. Recent work has proposed agentic formulations in which LLMs coordinate multi-step workflows, invoke tools, and interact with external knowledge sources, positioning such systems as emerging ``AI co-scientists'' rather than purely conversational interfaces \cite{autogen_2023,langgraph_2024,allenai_sciagents_2024}. In parallel, retrieval-augmented generation (RAG) has become a dominant strategy for grounding LLM outputs in domain-specific corpora, enabling interaction with large technical document collections while mitigating hallucination \cite{gao_rag_survey_2023,fan_ra_llm_survey_2024,zhao_rag_aigc_2024}.

Despite this progress, most existing agentic systems implicitly target cloud-centric and lightly governed environments. They often assume persistent internet access, homogeneous inference backends, ephemeral memory, and unrestricted sharing of retrieved knowledge across agents. These assumptions conflict with the operational realities of scientific computing in U.S.\ Department of Energy (DOE) laboratories, where researchers routinely work in restricted-network environments, rely on heterogeneous inference endpoints, and must maintain clear provenance and reproducibility of computational workflows \cite{doe_ai_science_2020,doe_ai_science_2023,whitehouse_aiforscience_2024}. As a result, many promising agentic frameworks remain difficult to deploy, extend, or sustain in production scientific settings.

This paper introduces \textbf{AISAC (AI Scientific Assistant Core)}, a modular framework designed to support LLM-driven scientific assistance under these constraints. AISAC is not positioned as an unconstrained autonomous AI scientist that independently defines research agendas or commits to externally consequential actions without human involvement. Instead, AISAC provides a reusable core architecture that enables \emph{contract-bounded autonomy}: given a user-initiated task, LLM agents can autonomously generate and evaluate hypotheses, decompose problems into structured plans, delegate subtasks to specialized research and review agents, invoke tools and retrieval systems, and compare competing lines of reasoning through controlled aggregation. These capabilities enable nontrivial scientific exploration while remaining constrained by runtime-enforced budgets, capability gates, and traceable execution boundaries. This distinction reflects established scientific accountability norms and institutional governance requirements.

A central motivation for AISAC is ease of adoption by the domain scientist---the combustion engineer, the materials scientist, the high-energy physicist, the climate modeler---who needs a capable, evidence-grounded AI assistant for their specific field but has no interest in building or maintaining AI infrastructure. Rather than embedding domain logic directly into orchestration code or requiring extensive reconfiguration for each new application, AISAC cleanly separates core execution infrastructure from project-specific customization. Scientists can define new agents, attach domain-specific retrieval corpora, and enable or disable tools through explicit registration and configuration hooks, without modifying the underlying framework. This design supports rapid spin-off of new scientific assistants from a shared core, enabling reuse across projects and research groups while preserving isolation between domains.

The primary contribution of this work is architectural rather than algorithmic. We describe how AISAC integrates multi-agent execution, persistent memory, and governed retrieval in a manner compatible with DOE-style workflows. In particular, AISAC treats agent-scoped retrieval and explicit extensibility as first-class concepts, allowing assistants to reflect domain structure while remaining easy to adapt and maintain. By foregrounding usability, reproducibility, and deployment realism, AISAC aims to bridge the gap between experimental agentic AI research and day-to-day scientific practice.

\paragraph{Paper Organization.}
Section~2 describes the design philosophy underlying AISAC and motivates its
focus on governed autonomy and structural design properties. Section~3 presents the
overall system architecture and routing model. Section~4 characterizes the structural properties of individual agents,
including capability scopes, execution profiles, and the execution graph. Section~5 details knowledge management and retrieval,
with emphasis on role-scoped access and explicit corpus lifecycle control.
Section~6 describes project customization via strict contracts that separate
core execution from project-owned extensions. Section~7 situates AISAC within related
work on agent frameworks, retrieval-augmented reasoning, and scientific AI
systems. Section~8 discusses limitations and future work, followed by
concluding remarks in Section~9.

\section{Design Philosophy}

The design of AISAC is guided by the principle that autonomy must be both useful and controllable. LLM agents drive internal reasoning and execution flow, but do so within boundaries explicitly defined at system initialization time: which models may be used, which tools are available, what knowledge sources exist, and how state is persisted. Agents cannot silently expand their capabilities or modify execution policy, an intentional departure from more self-modifying agent architectures \cite{reflexion_2023,camel_2023}.

AISAC distinguishes between execution autonomy and scientific responsibility. Once a task is initiated, agents may autonomously decompose problems, delegate subtasks, invoke tools, and interact with retrieval systems \cite{autogen_2023,langgraph_2024}. Externally consequential actions---job submission, data modification, index refresh---are not implicit runtime behaviors; they are expressed exclusively through explicitly registered project-level tools. This separation allows AISAC to support deployments ranging from tightly governed to more autonomous, while preserving traceability throughout.

Ease of extension and reuse is a first-class design concern. Scientists can register new agents, attach domain-specific retrieval corpora, and enable or disable tools through explicit configuration hooks, without modifying the shared core. Project-level customization is isolated behind a strict bootstrap mechanism, ensuring that extensions remain explicit, reversible, and auditable. This enables new scientific assistants to be spun off rapidly from a common core---a practical requirement in collaborative laboratory environments where groups share infrastructure but maintain distinct research agendas.

Finally, AISAC prioritizes structural clarity over maximal dynamism. Agent roles are explicitly defined, execution graphs are stable by default, and extensibility is mediated through well-defined interfaces. While this limits certain forms of agent self-modification, it substantially improves debuggability, predictability, and long-term maintainability---properties essential for scientific infrastructure intended to support sustained research programs rather than short-lived demonstrations.

\section{System Architecture}

\begin{figure}[!ht]
    \centering
    \includegraphics[width=\linewidth]{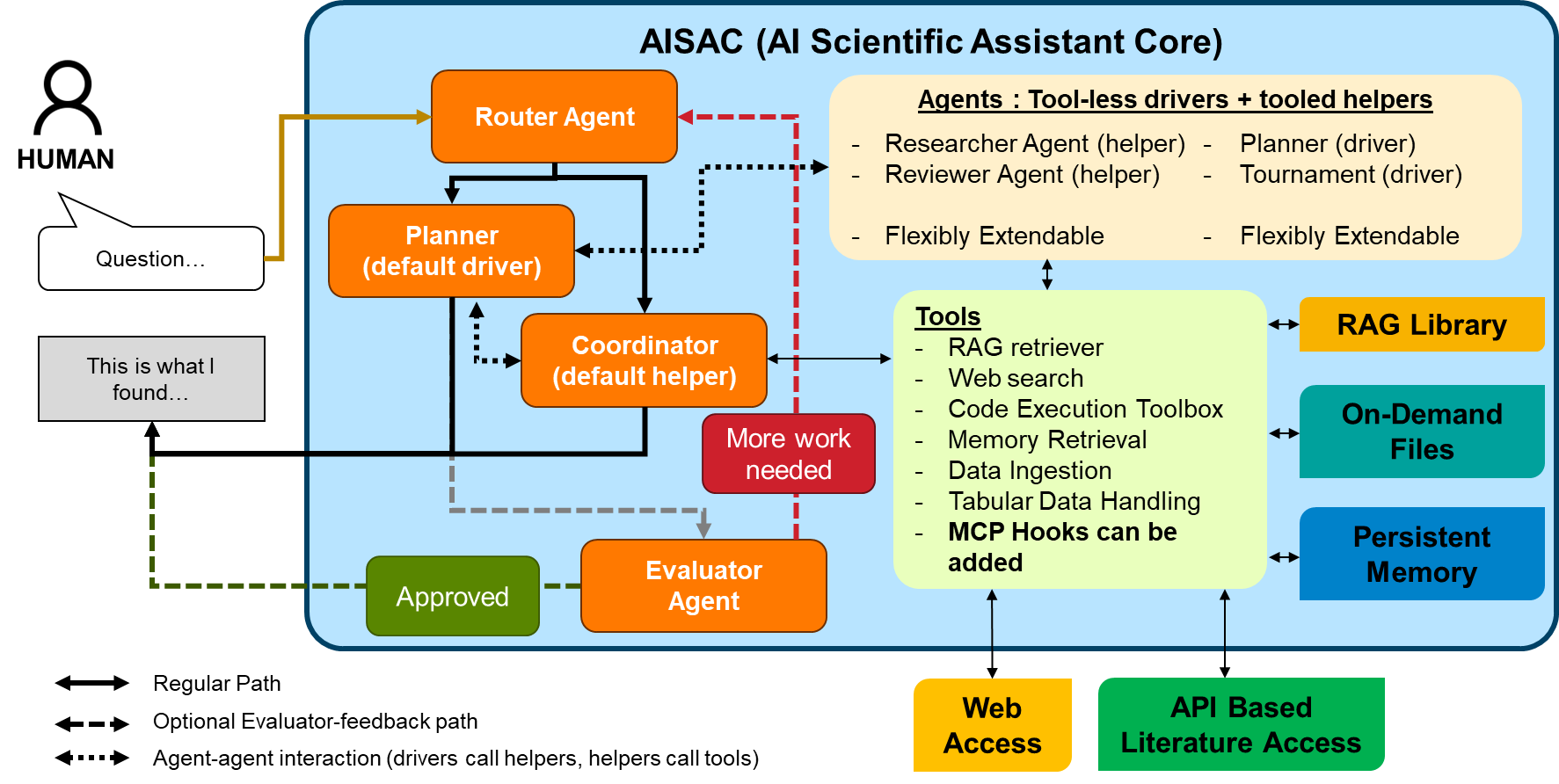}
    \caption{High-level component architecture of AISAC. Drivers (Router, Planner) handle task-level reasoning and delegation without invoking tools. Helpers (Coordinator, Researcher, Reviewer) execute bounded tool-mediated actions under explicit capability constraints. Hybrid persistent memory (SQLite + dual FAISS indices) and governed retrieval support long-horizon reasoning with full provenance.}
    \label{fig:aisac-basic-flow}
\end{figure}

Figure~\ref{fig:aisac-basic-flow} illustrates the overall schematic architecture of AISAC and the primary data and control flows between its components. The system is organized into a small number of stable layers that separate execution control, project customization, knowledge access, persistence, and deployment concerns. This layered structure is intended to support sustained scientific use, where clarity of responsibility, controlled execution depth, and explicit handoffs are more important than maximal dynamism.

\subsection{Routing and Agent Roles}

\begin{figure}[t]
  \centering
  \includegraphics[width=\linewidth]{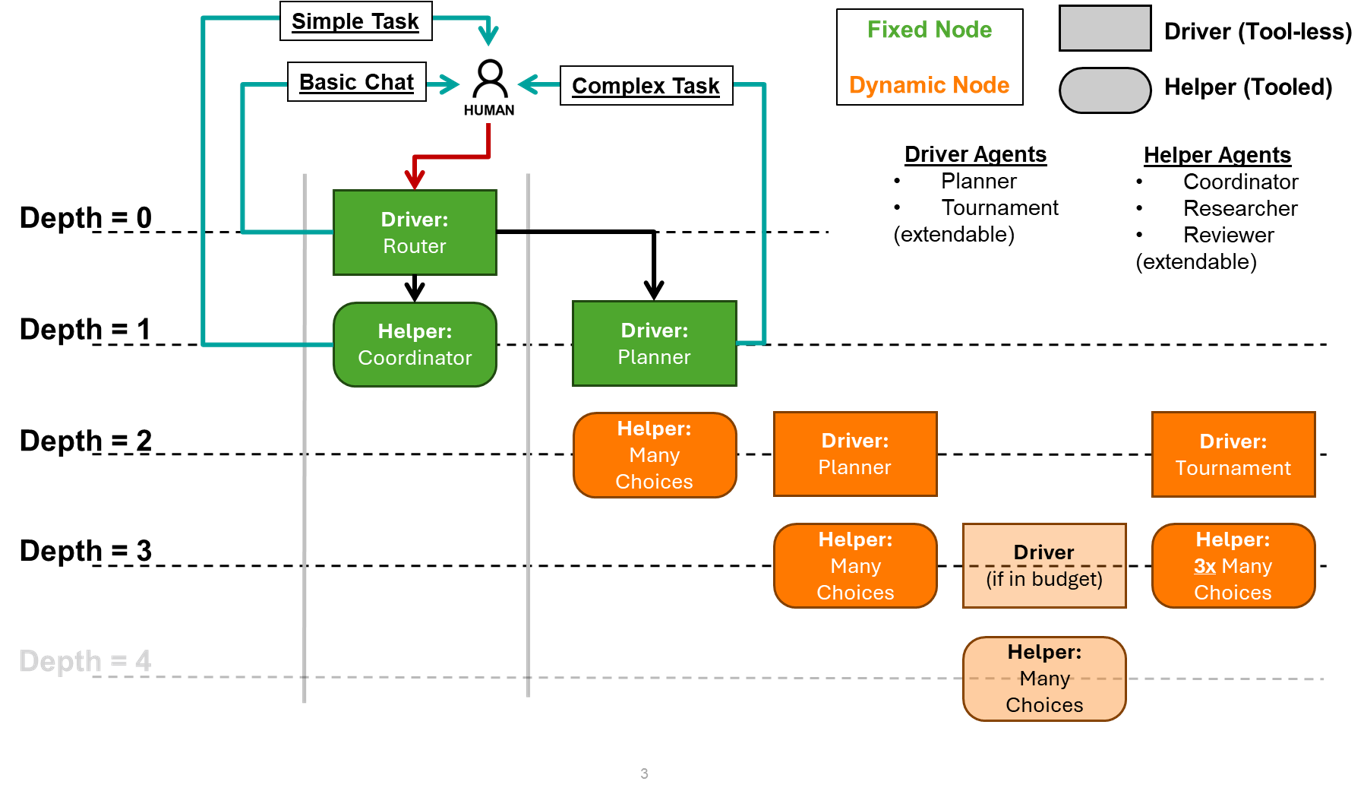}
  \caption{Execution flow and architectural layers of AISAC. User requests enter through the Router, which selects one of three paths: direct response, coordinator-mediated execution, or planner-driven multi-step reasoning. Project-specific agents, tools, and retrieval corpora enter through a strict bootstrap layer (Section~6). All execution is backed by hybrid persistent storage and routed through heterogeneous inference endpoints. \emph{Fixed nodes} (solid outlines) are always present in the graph; \emph{Dynamic nodes} (dashed outlines) denote positions whose active agent is resolved at runtime through state-based dispatch---all nodes are statically registered at session startup.}
  \label{fig:aisac_arch}
\end{figure}

At the entry point of the system is a \emph{router} (see Figure \ref{fig:aisac_arch}), which serves as the initial decision-making component for all user requests. The router is responsible for interpreting incoming queries, maintaining job continuity and conversational context, and selecting an appropriate execution pathway based on the complexity and intent of the request. Rather than treating all inputs uniformly, AISAC explicitly distinguishes between multiple routing outcomes, enabling selective invocation of agent depth only when warranted.  Rather than relying solely on truncation or summarization, the router selects specific historical dialogue turns for inclusion under the active context budget, enabling continuity while maintaining explicit control over long-horizon context growth. Routing decisions are constrained by the set of agents registered at runtime and by the remaining allowable delegation depth, preventing escalation to unavailable or disallowed execution paths. The router may additionally incorporate evaluator feedback from prior steps as an execution-control signal, enabling closed-loop governance without manual intervention.

The router supports three primary execution routes. First, for simple or well-scoped requests that can be addressed directly, the router may generate a response without engaging the agent execution pipeline. This path is intended for lightweight interactions that do not require external knowledge access, tool invocation, or multi-step reasoning.

Second, for requests that require limited retrieval, structured lookup, or straightforward tool-mediated execution, the router forwards control to the default \emph{coordinator} helper. This path enables bounded execution without explicit task decomposition and without access to the broader agent ecosystem reserved for planner-mediated execution.

Third, for requests that require multi-step reasoning, task decomposition, or specialized domain expertise, the router escalates execution to the default \emph{planner}, which is the default primary \emph{driver} agent. The planner is not a fundamentally different agent type, but a driver operating with expanded visibility and delegation authority. In particular, the planner has access to a broader set of specialized drivers (e.g., tournament or critique drivers) and helper agents that are not available to the lightweight execution path. This expanded access enables the planner to decompose complex tasks, select appropriate sub-agents, and explicitly coordinate multi-agent execution.

In AISAC, \emph{drivers} are responsible for task-level reasoning, planning, and synthesis, while \emph{helpers} are responsible for bounded execution. Drivers never invoke tools directly. Instead, drivers delegate all operational actions (e.g., retrieval queries or tool calls) to helpers, which execute those actions under explicit capability constraints. This separation enforces a clear boundary between reasoning and execution and prevents uncontrolled side effects during task coordination. Critically, this boundary is enforced \emph{architecturally} through the execution graph rather than by prompt convention: driver nodes in AISAC's state machine have no outgoing tool-call edges and are never exposed to tool invocation primitives \cite{langgraph_2024}. A driver agent cannot invoke a tool regardless of what its reasoning output contains, because the execution substrate does not provide that capability to driver-role nodes. This structural enforcement contrasts with frameworks where role separation is expressed as a prompt instruction and relies on agents voluntarily respecting declared boundaries---a design that is fragile under distribution shift and difficult to audit.

Execution proceeds through explicit, depth-bounded handoffs rather than unconstrained message passing. A planner may delegate subtasks not only to helpers but also to other specialized drivers, including additional planners, when deeper decomposition is required. Recursion depth is strictly bounded by runtime limits (by default, a maximum driver nesting depth of two levels beneath the router, configurable per deployment), ensuring that delegation cannot grow unbounded. At all depths, drivers retain coordination authority while helpers remain the sole agents permitted to execute tools. This structure enables expressive, multi-level task decomposition while preserving centralized control, traceability, and predictable execution behavior. Figure~\ref{fig:aisac_arch} illustrates the resulting execution structure: the router occupies depth zero; the coordinator (helper path) and planner (driver path) occupy depth one; and additional specialized drivers or helpers invoked by the planner occupy depth two. All graph nodes are registered statically at session startup; the active agent at each depth is resolved through state-based routing rather than dynamic node instantiation.

\subsection{Tooling, Capabilities, and Execution}

Tools in AISAC are treated as first-class execution primitives rather than incidental utilities. Tools encapsulate externally meaningful actions such as querying datasets, interacting with code repositories, or performing structured computations. Tools are never invoked directly by the user or implicitly by the system. All tool usage is mediated by agents and routed through the core execution layer, which enforces capability constraints and records tool invocations for provenance. This design ensures that externally consequential actions occur only within well-defined governance boundaries. The default tool catalog comprises over forty tools spanning several capability categories: literature and web retrieval (free API-based search across arXiv, PubMed, and Semantic Scholar, token-gated access to IEEE Xplore, Springer Nature, Elsevier/Scopus, and ScienceDirect when institutional keys are configured, and general web search; token-gated sources are silently excluded when keys are absent, preserving a functional tool for users without subscriptions), local semantic retrieval (hybrid BM25 lexical and dense vector search with score-normalized blending and per-source diversity controls), tabular data operations (loading, inspection, querying, and aggregation across CSV, Excel, Parquet, and JSON formats), persistent sandboxed code execution (stateful interpreter sessions with per-session variable persistence and a configurable scientific library allowlist enforced via AST-based static analysis), and document content extraction including table and image detection from scientific papers, reports, and presentations. Projects may extend this catalog through declarative tool registration without modifying the core.

\begin{figure}[t]
  \centering
  \includegraphics[width=\linewidth]{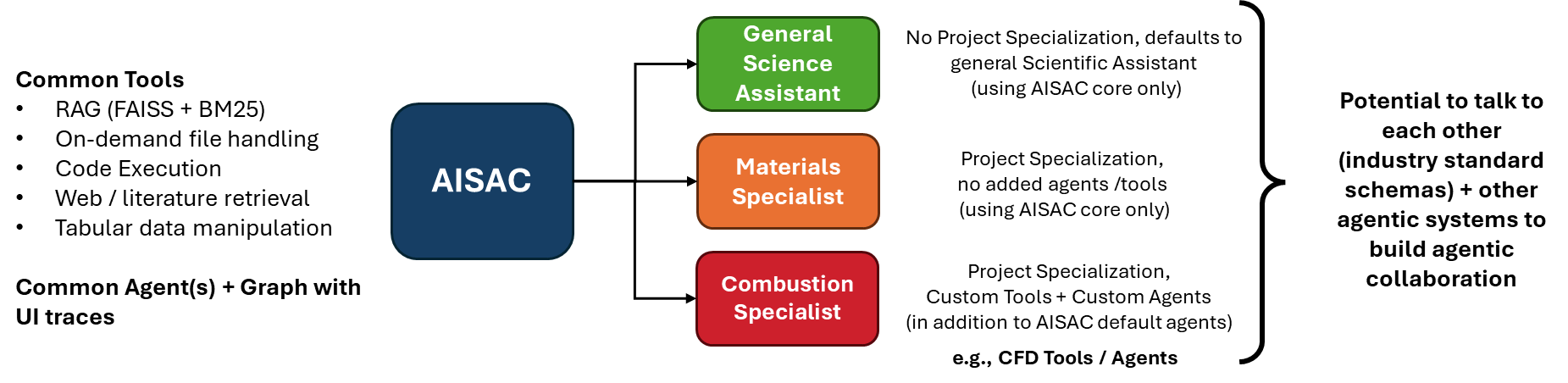}
  \caption{Project-level customization via strict bootstrap contracts. Out of the box, AISAC operates as a general-purpose scientific assistant with core tools and agents. Through fixed configuration hooks, projects can specialize AISAC with domain-specific retrieval corpora, custom tools, and tailored agent definitions without modifying the shared core.}
  \label{fig:bootstrap}
\end{figure}

Project-specific customization enters the system through a dedicated \emph{per-project adaptation layer}. This layer supplies agents, tools, retrieval roots, and high-level directives that tailor AISAC to a particular scientific application. As shown in Figure~\ref{fig:bootstrap}, all project inputs pass through a strict bootstrap mechanism that recognizes only a fixed set of integration points. Project components cannot modify core execution semantics; they can only populate predefined roles within the architecture. This contract-driven approach ensures that customization remains explicit and auditable while preserving stability of the core.

All LLM invocations and tool executions in AISAC are routed through standardized runtime invocation primitives rather than agent-defined call paths. These primitives enforce capability gates, attach execution metadata (agent identity, role, profile, endpoint, and delegation depth), and emit structured events consumed by the live event-stream interface. The interface provides real-time hierarchical visualization of agent delegation, tool calls, and intermediate outputs, along with live context panels that expose the working state of each active agent. A provenance inspector surfaces the detailed action log and complete execution trace for each session, serving as externalized working memory that is visible and auditable but not reinjected into agent prompts. This design centralizes governance, provenance, and UI observability in the runtime, ensuring that new agents inherit consistent execution behavior without reimplementing plumbing.

\subsection{Knowledge, State, and Persistence}

Knowledge access is mediated by the \emph{knowledge management layer}, which includes both global and agent-specific retrieval corpora. Helper agents may issue retrieval queries only against corpora explicitly assigned to them, and retrieved context is passed back to the invoking driver rather than being shared implicitly across agents. Retrieval lifecycle management is fully decoupled from execution: indices are constructed and refreshed only through explicit user action, never as a side effect of agent execution.

Persistent state is handled by the \emph{persistence and provenance layer}. At its core is an \emph{immutable execution trace}---a first-class architectural primitive rather than an incidental log. Every agent transition, delegation event, tool invocation, and retrieval result is recorded through a single centralized write path before any subsequent action proceeds, ensuring the trace is always consistent with actual execution state. The trace schema is fixed and complete by design: it captures not just outputs but agent identity, role, delegation depth, endpoint, and context budget at each step, enabling faithful replay and post-hoc audit. Beyond the trace, this layer records conversational context, intermediate outputs, and retrieval events with their execution metadata. Persistence supports inspection and traceability but does not imply automatic reuse. Historical memory is not injected into new executions unless explicitly requested, preventing hidden coupling between tasks or projects.

A complementary design choice concerns how agent reasoning is treated relative to tool execution. Consistent with the ReAct paradigm \cite{react_2023}, AISAC requires agents to articulate an explicit rationale alongside every tool invocation, expressed as a structured argument accompanying the tool call. This serves two purposes simultaneously. First, requiring a structured rationale before dispatch means the agent must commit to a stated reason before acting, which supports coherence across multi-step tool sequences. Because the rationale travels within the tool call record rather than as free-form content, it persists in the conversation history and remains visible to the agent on subsequent steps without occupying additional context. Second, the same rationale is extracted and surfaced through the live event-stream interface, making the agent's decision process directly observable to the supervising scientist at each step without requiring post-hoc log inspection. The same requirement applies to sub-agent invocations: when a driver delegates a subtask to a helper, it must supply an explicit rationale for that delegation alongside the task description. As a result, every consequential LLM action in AISAC---whether a tool call, a retrieval query, or a delegation to a specialist agent---carries a stated justification, creating a complete and human-readable record of intent across the full execution trace.

In addition to the append-only execution trace, AISAC provides a \emph{shared agent workspace} (blackboard) for mutable collaborative state. The blackboard is a database-backed store scoped by user, task, and named slot, through which agents can write, read, append, patch, and search shared content during execution. This enables multi-step collaborative artifacts---such as accumulating analysis sections, building structured reports, or tracking intermediate conclusions across successive helper invocations---without inflating agent prompts. All writes are attributed to the authoring agent and timestamped, preserving provenance for mutable working state distinct from the immutable execution record. Concurrent writes to the same slot are serialized by the database write layer; agents requiring coordinated multi-step updates should use the append or patch operations rather than overwriting, allowing contributions from multiple helpers to accumulate without collision.

This design deliberately revives the classical \emph{blackboard architecture}---a shared, structured problem space through which heterogeneous specialist agents coordinate without requiring direct peer-to-peer communication \cite{hayes_roth_blackboard_1985}. The pattern, central to early AI systems such as HEARSAY-II, has been largely displaced in LLM-based multi-agent frameworks by conversational message-passing (e.g., agent-to-agent chat turns \cite{autogen_2023}), which couples agents through growing conversation histories and inflates context as collaboration deepens. AISAC's blackboard decouples agents from each other's internal reasoning context while preserving shared artifact state, enabling scalable multi-agent collaboration without prompt inflation or implicit knowledge leakage between agents. The attribution and timestamp metadata on all writes extend the classical pattern with provenance semantics appropriate for accountable scientific workflows.

\subsection{Deployment, Federation, and Runtime Control}

Finally, inference is provided by a dedicated \emph{deployment and endpoint layer}. AISAC supports multiple heterogeneous inference endpoints concurrently and allows different agents to utilize different models within the active endpoint, while decoupling chat and embedding inference across independent backends. Endpoint selection is resolved outside agent logic through configuration and execution policy, ensuring that changes in deployment environment do not require modifications to project code or agent definitions. For DOE-available inference services, including Argonne-internal endpoints, AISAC integrates institutional authentication flows with automatic token refresh, enabling non-interactive access to restricted backends.

AISAC also integrates with the Model Context Protocol (MCP) to incorporate external partner agents running on separate infrastructure, including HPC login nodes and remote services. Partner agents are discovered at startup via a server-sent events (SSE) bridge, with tool schemas extracted from machine-readable definitions or inferred from structured docstrings. Once discovered, external agents are registered as helpers or tools within the existing execution model, inheriting the same capability gates, provenance tracking, and delegation semantics as native agents. Access to partner agents is mediated through institutional group membership (e.g., Globus groups), enabling fine-grained sharing of computational resources and specialized agents across project boundaries without public exposure. This allows, for example, a researcher's workstation-hosted AISAC instance to transparently invoke HPC-resident analysis agents for compute-intensive subtasks, with the MCP bridge handling authentication, serialization, and result return within the existing driver-to-helper delegation model. Federated trust across institutional boundaries remains a partially open problem; AISAC's current approach addresses access control through group membership but does not provide formal attestation of partner agent behavior or output provenance. The design nonetheless enables federated scientific workflows that span institutional boundaries without requiring modifications to core execution semantics.

AISAC provides first-class deployment support across the environments in which scientific work occurs. A desktop launcher application manages multiple concurrent project instances with independent configurations, credential stores, and port assignments, presenting each project as a native application window on Windows, macOS, and Linux. A project setup wizard guides scientists through endpoint selection, credential configuration, retrieval corpus registration, and feature enablement without requiring manual editing of configuration files. For HPC environments, AISAC provides terminal-based launchers that integrate with module systems and network proxies, with the Gradio interface accessed via SSH port forwarding; institutional authentication tokens are obtained via browser-based PKCE flows and refreshed automatically in the background for non-interactive operation. An offline mode disables all external network tools while preserving local retrieval, code execution, and tabular analysis capabilities, supporting air-gapped or restricted-network deployments. Across all environments, project configuration is expressed as a single bound object loaded at process start, ensuring that deployment environment does not affect core execution semantics or agent behavior.

AISAC provides runtime execution control mechanisms that give users direct oversight of active multi-agent workflows. Three runtime signals are available: a \emph{stop} signal that halts execution immediately, an \emph{answer-now} signal that cascades through nested drivers and forces each to summarize its partial results before returning, and an \emph{interject} signal that injects a user note into the active agent's context without interrupting execution. In addition, driver and helper agents may request user clarification mid-execution. Clarification requests carry configurable severity levels: low-severity requests auto-resolve after a timeout, allowing execution to proceed with the agent's best judgment, while high-severity requests block until the user responds. Per-agent clarification budgets prevent agents from stalling execution with excessive queries. Together, these mechanisms operationalize bounded autonomy as a runtime property rather than a static policy.

Agents also operate under explicit \emph{context budget awareness}. Each agent receives a runtime banner reporting current context utilization relative to the model's capacity, and agent instructions encode graduated behavioral policies tied to configurable budget thresholds: reducing retrieval breadth as context fills, suppressing new tool invocations as capacity is approached, and finalizing output before the window is exhausted. Sensible defaults are provided for each threshold and can be adjusted at the project level when deploying against endpoints with larger or smaller context windows, without modifying agent logic. Per-turn execution is likewise bounded by configurable limits on total tool invocations and conversation turns, with soft warning levels that give agents the opportunity to conclude gracefully before hard caps are enforced. Together, these defaults ensure that long-running tasks remain within predictable resource envelopes and degrade gracefully rather than truncating silently.

Taken together, these architectural layers define a system in which responsibility is clearly partitioned between drivers and helpers, and execution depth is selectively invoked through routing rather than emergent behavior. The router enables efficient handling of simple requests, while planner-mediated execution allows controlled access to richer agent ecosystems when deeper reasoning is required. This architecture allows AISAC to deliver practical scientific assistance while preserving the control, transparency, and predictability required for DOE scientific environments.

\section{Execution Model and Agent Roles}

AISAC adopts an explicit agent-based execution model in which responsibility for
reasoning, coordination, and domain expertise is partitioned across a small
number of well-defined agent roles. This design is motivated by two practical
considerations. First, scientific workflows naturally involve distinct cognitive
functions, such as problem decomposition, hypothesis generation, critique, and
synthesis of results. Second, clarity in agent responsibilities simplifies both
system behavior and user understanding, reducing the cognitive overhead required
to interpret, debug, or audit agent execution. AISAC therefore favors a small set
of stable agent roles over dynamically evolving or self-modifying agent
populations.

Section~3 describes the routing model, driver--helper separation, and depth-bounded delegation structure that govern task flow. This section focuses on the structural properties of individual agents within that model: capability scopes, execution profiles, and the execution graph.

Helper agents operate within tightly defined capability scopes. Each helper may
be granted access to a specific subset of tools, retrieval corpora, or external
resources, reflecting its intended role. AISAC supports agent-specific retrieval
scopes, allowing individual helpers to query curated knowledge bases aligned with
their domain expertise. Retrieved information and tool outputs are returned to
the invoking driver in structured form, enabling coherent synthesis while
preserving centralized control and interpretability.

AISAC further decouples agent identity from model instantiation through the concept of agent execution profiles. A single logical agent may expose multiple profiles, each binding a specific language model, prompt configuration, or decoding policy (e.g., code-oriented synthesis versus long-context reasoning). Calling drivers may explicitly select which profile of a callee to invoke based on task needs, and profile selection is recorded as part of the execution trace.

Profiles may additionally be scoped to specific inference endpoints, allowing the same logical agent to exhibit different execution behavior across heterogeneous backends while preserving stable agent roles and delegation structure.

AISAC’s execution model emphasizes determinism and transparency over maximal
flexibility. The sequence of agent interactions is governed by an explicit
execution graph rather than unconstrained message passing. While the default
execution graph is stable, advanced users may extend or override it through a
controlled bootstrap mechanism. This approach balances extensibility with
predictability, allowing new agents or execution paths to be introduced without
sacrificing debuggability or reproducibility, which are critical in scientific
contexts \cite{doe_ai_science_2023}.

Throughout execution, AISAC maintains a persistent record of agent interactions,
delegation events, tool invocations, and intermediate outputs. These records are
stored in a database-backed memory system and are available for inspection or
replay. Importantly, persistence does not imply unbounded reuse of prior context.
Historical information is not automatically promoted into active reasoning
contexts without explicit user intent, preventing hidden coupling across tasks
or projects and avoiding uncontrolled context growth \cite{memgpt_2023}. Beyond the execution trace, AISAC supports two forms of cross-session state managed through explicit agent tools: \emph{agent memory}, in which individual agents may persist durable notes (key-value pairs, database-backed) that carry forward across sessions, enabling agents to accumulate domain-specific observations without inflating prompts; and \emph{user preferences}, a per-user record that is auto-injected into agent contexts, allowing agents to adapt to established conventions and workflows over time.

From the scientist's perspective, this execution model is intentionally transparent without being burdensome. Scientists interact at the level of task specification and result interpretation rather than agent choreography; the underlying delegation structure, tool usage, and retrieval decisions are visible in the live trace but require no manual intervention during normal operation.

\section{Knowledge Management and Retrieval}

AISAC treats knowledge access as a first-class, governed system concern rather than an incidental feature of agent execution. In scientific environments, the provenance, scope, and lifecycle of knowledge sources are often as important as the reasoning mechanisms that consume them. Consequently, AISAC avoids designs in which agents implicitly ingest data from uncontrolled sources or silently accumulate context across tasks. Instead, all retrieval capabilities are explicitly configured, scoped, and managed by the user, aligning with DOE expectations for transparency and reproducibility \cite{doe_ai_science_2023,whitehouse_aiforscience_2024}.

Retrieval-augmented generation (RAG) in AISAC is implemented through a clear separation between \emph{global} and \emph{agent-specific} retrieval corpora. Global corpora represent knowledge that is broadly applicable across a project, such as shared documentation, standards, or foundational references. In contrast, agent-specific corpora are attached to individual agents and are intended to reflect narrowly scoped domain expertise, such as a particular codebase, experimental dataset, or set of technical manuals. This design mirrors the structure of real scientific teams, where expertise is intentionally partitioned rather than universally shared, and enables agents to reason deeply within their intended domains without contaminating other agents with irrelevant or out-of-scope context \cite{gao_rag_survey_2023,graphrag_2024,paperqa_2023}.

AISAC’s execution model enforces these retrieval boundaries by construction. Each helper agent is assigned a retrieval scope that may include the project-wide global corpus, a private agent-specific corpus, or both. An agent configured with a private corpus can draw on specialized knowledge---a domain-specific dataset, a curated set of technical references, a project codebase---in addition to the shared global corpus, without exposing that knowledge to other agents or contaminating their retrieval context. Retrieved context is returned to the coordinating driver agent in structured form, insulating it from low-level retrieval details while preserving visibility into provenance and relevance. This approach contrasts with monolithic retrieval setups in which all agents share a single global knowledge pool, often leading to uncontrolled context growth and reduced interpretability.

The lifecycle of retrieval corpora in AISAC is intentionally conservative. Retrieval indices are never built, refreshed, or expanded implicitly at system startup or during agent execution. Instead, corpus construction and refresh are treated as explicit user actions, ensuring that data ingestion events are deliberate, auditable, and computationally predictable. This policy avoids unintended mixing of datasets, unclear temporal provenance, and unnecessary recomputation, all of which are common sources of error in long-lived scientific assistants. By decoupling retrieval lifecycle management from agent execution, AISAC enables scientists to reason confidently about what knowledge is available to the system at any point in time. Content eligibility for indexing is governed by explicit policy (e.g., on-demand artifacts are never indexed), rather than inferred solely from filesystem location or file type.

Beyond explicit user control over when indices are built or refreshed, AISAC also treats \emph{indexability itself} as a policy decision rather than an implementation detail. Certain classes of content—such as large tabular artifacts, slide decks, or ephemeral analysis outputs—are designated as on-demand resources and are explicitly excluded from indexing, even if they reside within retrieval paths. This distinction prevents uncontrolled growth of retrieval indices, avoids semantic dilution from low-signal artifacts, and preserves clear provenance boundaries between curated evidence and transient working materials.

Retrieval in AISAC is shaped not only by the choice of dense and lexical retrieval methods, but also by explicit aggregation and filtering policies applied after candidate generation. Retrieved results may be subject to per-source or per-document caps, optional diversity constraints, and agent-scoped supersampling rules that prioritize evidence from designated knowledge roots. These mechanisms ensure that downstream reasoning operates over a bounded, interpretable set of evidence rather than an unstructured accumulation of retrieved text, supporting both long-horizon reasoning stability and traceable evidence use.

AISAC’s knowledge management design also supports ease of extension and project spin-off. New scientific assistants can be created by defining new retrieval corpora and associating them with existing or newly defined agents, without modifying the core execution infrastructure. This enables rapid adaptation to new domains or research questions while preserving a stable, shared core. In practice, this allows research groups to maintain multiple assistants tailored to different projects or campaigns, each operating over its own curated knowledge base while benefiting from the same underlying framework.

By combining explicit retrieval scoping, conservative lifecycle management, and tight integration with the agent execution model, AISAC provides a knowledge management approach that is both powerful and predictable. The system enables LLM agents to leverage rich domain knowledge where appropriate, while preserving the transparency, reproducibility, and ease of use required for sustained deployment in DOE scientific environments. Reproducibility in this context refers specifically to the deterministic behavior of the retrieval, tool, and trace layers: given the same corpus and configuration, the same retrieval paths, tool invocations, and execution records will be produced. Full end-to-end reproducibility of responses additionally requires controls on LLM inference such as temperature pinning and model version locking, which AISAC supports at the configuration level but does not mandate.

\section{Project Customization via Strict Contracts}

AISAC is designed as a reusable core rather than a single monolithic application. In practice, scientific groups require assistants that differ in their tools, agent expertise, retrieval corpora, and operating constraints, yet they also benefit from a common, stable execution substrate. AISAC addresses this need by separating \emph{project-owned customization} from \emph{core execution infrastructure}, and by funneling all customization through strict, explicit contracts. The purpose of these contracts is twofold: to reduce the engineering burden on domain scientists and to preserve predictability, debuggability, and governance across deployments.

A project in AISAC is defined by a small set of fixed integration points, each of which is optional and controlled through explicit configuration toggles. Rather than allowing arbitrary module discovery or dynamic plugin loading, AISAC uses a strict bootstrap mechanism that recognizes only a fixed set of project-owned files and a constrained set of hooks. This design prevents import-time side effects, avoids accidental capability escalation, and makes it straightforward for scientists to understand what parts of the system are project-specific versus core.

The primary project-owned configuration is provided through a bound project configuration object. This configuration defines project identity and high-level operating constraints (e.g., offline operation), as well as the locations of curated retrieval corpora. The core binds this configuration at process start and treats it as read-only state for the duration of execution. Importantly, project configuration specifies \emph{what knowledge exists} (e.g., global versus on-demand corpora), but it does not cause any implicit ingestion or indexing. Knowledge lifecycle actions remain explicit user choices, consistent with AISAC's governance-first design.

Project-specific tools are integrated through a single, narrow interface: a project may provide a function that returns a list of tool definitions to be registered into AISAC's tool catalog. Tools are therefore added declaratively rather than through ad hoc mutation of core logic. This keeps the extension surface small, supports incremental adoption, and makes it easy to audit which tools are available in a given project deployment.

Project-specific agents are integrated through a similarly constrained mechanism. A project may register agents either through a designated registration hook or through declarative agent specifications, both of which are ultimately expressed through a single core registration contract. This contract captures not only the agent's identity and system prompt, but also explicit capability gates and governance controls. AISAC supports fine-grained per-agent tool access: rather than a binary grant of the full tool catalog, each agent is assigned a named set of tool families (e.g., retrieval, code execution, web search, data analysis, file operations), allowing capability scope to be matched precisely to the agent's intended role. In addition, per-agent restrictions on web access and global retrieval access can be set independently, and per-agent retrieval roots define each agent's private knowledge scope. These capability assignments are surfaced in the project configuration and can be inspected and adjusted through the graphical interface without modifying code, making it straightforward for domain scientists to tailor agent behavior incrementally. Together, these features allow projects to define assistants in which expertise and access are deliberately partitioned across agents, with knowledge and tool boundaries enforced by construction rather than by convention.

Finally, AISAC permits controlled customization of execution structure. The default execution graph is stable and shared across projects, but advanced users may override the graph construction through a single explicit hook. This provides a pragmatic balance: most projects can adopt AISAC as-is, while projects with specialized requirements can extend execution behavior without forking the core. Because such overrides are explicit and localized, they remain inspectable and maintainable over time.

To lower the barrier further for scientists who prefer not to edit configuration files directly, AISAC provides a desktop launcher and project setup wizard that guide users through endpoint selection, credential configuration, retrieval corpus registration, and feature enablement via a graphical interface, generating a well-formed project directory as output. Once a project is running, the Gradio-based interface exposes agent execution traces, live context panels, and runtime controls through the same browser window, making the system inspectable and adjustable without developer intervention.

The primary target user is the domain scientist---a combustion researcher, a materials scientist, an energy systems analyst---who wants a capable, evidence-grounded AI assistant for their specific field but has no interest in building or maintaining AI infrastructure. AISAC is designed with the goal that a scientist can configure it once for their domain and then use it as a daily research collaborator, without ongoing framework maintenance---analogous to how domain-specific professional tools are adopted and used, but purpose-built for agentic scientific reasoning. The key difference is that AISAC is purpose-built for scientific workflows, with the governed retrieval, traceable execution, and heterogeneous inference support that general-purpose consumer applications do not provide. A research group can spin up multiple domain-specialized AI co-scientists from the same core---a combustion science assistant, a materials characterization assistant, a process safety assistant---each acting as a knowledgeable collaborator for its domain, with its own curated corpus, project-level agents, and tool access, without duplicating infrastructure or requiring AI expertise beyond the initial configuration.

Taken together, these strict contracts define a project layer that is expressive enough to capture real scientific customization---tools, agents, directives, and retrieval scopes---while remaining small enough for domain scientists to adopt without becoming framework developers. This contract-first approach is a central mechanism by which AISAC achieves both ease of use and sustainable governance in DOE scientific environments.

\section{Related Work}

Recent years have seen rapid growth in agentic language model systems that extend large language models beyond single-turn interaction toward multi-step reasoning, tool use, and collaborative execution. Frameworks such as AutoGen and LangGraph provide general-purpose abstractions for constructing agent workflows, message passing, and execution graphs \cite{autogen_2023,langgraph_2024}. These systems emphasize flexibility and developer expressiveness, enabling rapid experimentation with agent behaviors and coordination patterns. AISAC is implemented on top of LangGraph's StateGraph runtime \cite{langgraph_2024}, using it as an execution substrate while adding governance contracts, runtime-enforced role semantics, budgeted context management, and structured provenance recording as layers that LangGraph itself does not provide. AISAC is therefore complementary to LangGraph rather than competitive with it: it narrows LangGraph's general-purpose graph model into an opinionated scientific assistant runtime where the trade-off between flexibility and governance is deliberately resolved in favor of the latter.

Several systems position agents as emerging ``AI co-scientists,'' focusing on autonomous hypothesis generation, literature exploration, or end-to-end research workflows. Examples include SciAgents and recent conceptual proposals for autonomous scientific discovery \cite{allenai_sciagents_2024,deepmind_co_scientist_2025}. While these systems explore the frontier of agent autonomy, they often assume permissive execution environments and emphasize agent intelligence over system control. In contrast, AISAC deliberately constrains autonomy, prioritizing explicit execution boundaries, traceability, and human oversight to align with institutional scientific practice and accountability requirements.

Retrieval-augmented generation (RAG) has become a foundational technique for grounding language models in external knowledge, with extensive work exploring dense, sparse, and hybrid retrieval strategies \cite{gao_rag_survey_2023,fan_ra_llm_survey_2024,zhao_rag_aigc_2024}. Systems such as GraphRAG and PaperQA demonstrate how structured or document-centric retrieval can improve reasoning quality in technical domains \cite{graphrag_2024,paperqa_2023}. AISAC builds on these insights but treats retrieval not as a monolithic service, but as a governed resource with explicit scoping, lifecycle management, and agent-specific access controls.

Persistent memory for agent systems has also received increasing attention, particularly in the context of long-running interactions and context management \cite{memgpt_2023}. Many approaches focus on optimizing what information should be retained or recalled to improve agent performance. AISAC adopts a complementary perspective, emphasizing separation between memory and knowledge, explicit persistence boundaries, and provenance-aware storage to support reproducibility and auditability in scientific workflows rather than purely improving conversational continuity. 

\begin{table}[H]
\centering
\footnotesize
\setlength{\tabcolsep}{3pt}
\renewcommand{\arraystretch}{1.12}
\begin{tabular}{p{2.7cm} p{3.4cm} p{2.6cm} p{2.6cm} p{2.6cm}}
\toprule
\textbf{Dimension} &
\textbf{AISAC (this work)} &
\textbf{LangGraph} &
\textbf{AutoGen} &
\textbf{MemGPT} \\
\midrule

Primary focus &
Scientific trustworthiness: provenance, reproducibility, and evidence-grounded workflows &
Stateful agent orchestration &
Conversational multi-agent interaction &
Long-horizon memory for a single agent \\

Architectural role &
Opinionated scientific-assistant \emph{runtime} with enforced execution contracts &
General-purpose execution substrate &
Agent interaction framework &
Memory-centric cognitive architecture \\

Execution model &
Bounded state graph with explicit depth and per-turn context budgets &
StateGraph with conditional and cyclic edges &
Turn-based conversational loops &
Continuous loop with memory paging \\

Agent role semantics &
\textbf{Explicit, runtime-enforced driver--helper roles} &
Roles encoded structurally but not semantically enforced &
Roles expressed primarily via prompts and system messages &
Single agent; no role separation \\

Tool usage discipline &
\textbf{Capability-gated:} drivers delegate to helpers only; tools executed by helpers with schema validation and logging &
Tool nodes allowed; governance user-defined &
Tools callable by agents; minimal enforcement &
Tools support memory and external actions \\

Memory, context, and horizon &
\textbf{Budgeted long-horizon reasoning:} router-managed context continuity, explicit depth limits, role-scoped hybrid memory (SQLite execution traces + dual FAISS for dialogue and evidence RAG), shared agent workspace (blackboard), and cross-session agent memory &
State persistence via checkpoints; context management user-defined &
Chat history with optional persistence &
Unbounded long-horizon memory via paging mechanisms \\

Transparency \& provenance &
\textbf{Built-in:} live event-stream GUI, replayable execution traces, and structured provenance &
Graph and state inspection (developer-facing) &
Conversational logs &
Memory access visibility \\

Customization model &
\textbf{Declarative bootstrapping} with closed, boolean-gated extension surfaces enabling structural auditability across projects &
Programmatic graph construction &
Prompt and agent configuration &
Memory policy tuning \\

Best suited for &
Long-horizon scientific workflows, lab automation, and auditable analysis &
Complex adaptive workflows and experimentation &
Collaborative coding and discussion-driven tasks &
Persistent personal assistants \\

Tradeoffs &
Reduced flexibility in favor of governance, reproducibility, and bounded resource usage &
Governance left to application logic &
Weaker provenance and lifecycle guarantees &
Limited multi-agent reasoning \\

\bottomrule
\end{tabular}
\caption{Comparison of AISAC with representative agent frameworks. AISAC emphasizes governed long-horizon reasoning via budgeted context management, runtime-enforced role semantics, and explicit provenance, while other frameworks prioritize orchestration flexibility, conversational interaction, or memory scalability.}
\label{tab:framework_comparison}
\end{table}

Crucially, AISAC's execution trace is not a logging side effect but a designed architectural primitive: written through a single centralized path, carrying a fixed schema that includes agent identity, role, delegation depth, and context budget, and guaranteed complete and consistent with actual execution state. This positions the trace as an instrument of scientific reproducibility---analogous to a structured lab notebook entry---rather than a developer debugging artifact, and enables post-hoc audit and replay in a manner that memory-paging approaches do not directly support.

Finally, a growing body of policy and programmatic work has articulated requirements for trustworthy, reproducible AI in scientific research, particularly within the DOE and broader U.S.\ federal research ecosystem \cite{doe_ai_science_2020,doe_ai_science_2023,whitehouse_aiforscience_2024}. AISAC directly operationalizes these requirements at the runtime level, translating high-level principles—such as transparency, provenance, and governance—into concrete architectural constraints. In this sense, AISAC should be viewed not as a competitor to existing agent frameworks, but as an execution substrate that enables such systems to be deployed responsibly in real-world scientific environments.

\section{Limitations and Future Work}

AISAC is presented as a governed execution substrate rather than a complete
scientific agent. While this positioning enables clarity, extensibility, and
deployment realism, it also introduces limitations that motivate future work.

\textbf{Evaluation scope.} This paper primarily contributes architecture and runtime-verified design properties. AISAC is in active operational use across multiple scientific projects at Argonne, and task-level evaluation across those deployments is ongoing; detailed results are not included in this version of the manuscript and will be reported separately. The current emphasis is therefore on the structural properties of the runtime: role semantics, traceability, bounded context management, and reproducibility at the tool and retrieval layers. Future reporting will address systematic end-to-end evaluation across representative scientific workloads (e.g., literature triage, code-assisted analysis, experimental planning, and simulation post-processing), measuring task success, evidence-groundedness, latency, and user time saved, along with ablations over routing choices, depth limits, and agent composition.

\textbf{Policy versus mechanism.} AISAC intentionally separates capability
definition from execution policy. Externally consequential actions (e.g., job
submission, data modification, or index refresh) are not implicit behaviors; they
are expressed only via explicitly registered project-level tools and agents, and
their availability is a deployment choice. This makes AISAC flexible, but it also
means safety and governance posture are ultimately project-dependent. Future work
includes providing reusable, formally specified policy modules (e.g., explicit
approval gates, immutable-data modes, or sandboxed execution profiles) that can
be composed with projects without modifying core.

\textbf{Long-horizon memory quality.} AISAC provides budgeted long-horizon
operation via hybrid persistence and role-scoped memory access, but it does not
solve the open problem of selecting optimal long-term memories for scientific
reasoning. In particular, storing and retrieving information is not equivalent
to maintaining a coherent evolving scientific model. Future work includes richer
memory compaction and consolidation strategies, provenance-aware summarization,
and explicit mechanisms for representing and revising hypotheses, assumptions,
and intermediate conclusions over multi-day workflows.

\textbf{Retrieval lifecycle and corpus drift.} AISAC intentionally decouples RAG
indexing from agent execution and treats corpus refresh as an explicit action to
preserve provenance and computational predictability. This conservative posture
can be inconvenient in settings where corpora change frequently. The current
implementation already supports content-hash-based incremental refresh (SHA-256
per file, diffed against a per-store manifest) and records index-level provenance
including the embedding endpoint and model used to construct each index. Remaining
future work includes dataset version pinning and optional scheduled refresh
policies that remain auditable and reversible.

\textbf{Tooling robustness and sandboxing.} AISAC currently enforces a code execution sandbox that combines AST-based static analysis with a module allowlist, blocking unsafe operations (e.g., filesystem access, subprocess invocation, network calls) while permitting scientific computation libraries. AST-based analysis operates on the static structure of submitted code and does not catch all dynamic execution patterns; constructs that generate or evaluate code at runtime fall outside its detection scope. Robust tool use in scientific environments also requires fault isolation, partial failure recovery, resource quotas, and containerized execution. Future work includes strengthening the sandbox boundary, adding per-call resource limits, and standardized error-recovery protocols that preserve trace integrity under failures.

\textbf{Heterogeneous inference endpoints.} AISAC is designed to operate across
multiple inference backends, enabling practical deployment under changing
infrastructure constraints. However, endpoint heterogeneity introduces new
systems challenges: latency variance, capability mismatch (e.g., tool-call
support), and reproducibility across model versions. Future work includes
explicit endpoint capability descriptors, execution policies that optimize for
cost/latency under constraints, and model-version pinning and reporting to
improve replayability of traces.

\textbf{From substrate to autonomous scientific systems.} AISAC is intentionally
a capable but incomplete skeleton. Its architecture can support a spectrum of
autonomy levels by adding specialized agents (e.g., hypothesis proposer,
experimental designer, critic/reviewer, or tournament aggregators) and by
relaxing or tightening execution policies. A key research direction is to
characterize which autonomy regimes are beneficial for scientific work while
remaining auditable and controllable, and to identify principled interfaces for
human oversight that do not collapse into manual agent choreography.

Overall, these limitations motivate future work that strengthens AISAC along two
axes: (i) quantitative evaluation on real scientific workloads, and (ii) richer
policy and reliability mechanisms that preserve AISAC’s core invariants while
supporting increasingly autonomous, domain-specialized scientific assistants.

\section{Conclusion}

This paper introduced \textbf{AISAC (AI Scientific Assistant Core)}, a governed multi-agent runtime that contributes a systems-level execution substrate for deploying agentic AI in scientific practice. Its primary design properties---runtime-enforced role semantics, budgeted context management, agent-scoped retrieval, and structured provenance---are motivated by the operational realities of DOE laboratory environments and the practical needs of domain scientists who require capable AI assistance without AI infrastructure expertise.

A central architectural principle is the separation of \emph{mechanism} from \emph{policy}. The core runtime provides expressive primitives for multi-agent reasoning, hierarchical delegation, and tool-mediated action, while project-level configuration determines autonomy regimes, available tools, and knowledge scope. This allows AISAC to serve as a reusable substrate across a wide range of domain-specialized assistants---from tightly scoped analysis tools to more exploratory, hypothesis-driven collaborators---without modification of the shared core.

AISAC is intended not as a final form of an autonomous AI scientist, but as a stable foundation on which diverse scientific assistants can be constructed, evaluated, and evolved. As agentic AI systems continue to mature, architectures that foreground transparency, bounded autonomy, and explicit control are likely to play a central role in their sustained use in scientific research. AISAC represents one step toward such architectures.

\vspace{1em}
\noindent\textbf{Acknowledgements:}
The authors acknowledge funding support from Argonne's Laboratory Directed Research and Development (LDRD) program via LDRD2025-0486 and LDRD2026-0214. Additional support was provided by the U.S.\ Department of Transportation's Bureau of Transportation Statistics through the AI for Energy Process Safety (AI-EPS) research program, and by the Genesis AI mission's COMB-FLOW project. The authors thank the project teams for their engagement and feedback during the development of this work.

\section*{Code and Data Availability}
Code Availability: AISAC is under active development and in operational use across multiple scientific projects at Argonne National Laboratory. The architecture and design properties described in this report reflect the current state of the system; capabilities continue to evolve as new domains and deployment requirements are addressed. We are evaluating pathways for broader release and welcome feedback on this technical report.

\bibliographystyle{unsrt}
\bibliography{references}

\end{document}